\documentclass[conference]{IEEEtran}
\IEEEoverridecommandlockouts

\usepackage{cite}

\usepackage{amsmath, amssymb, amsfonts}

\usepackage{graphicx}

\usepackage{textcomp}
\usepackage{xcolor}

\usepackage{array}
\usepackage{multirow}
\usepackage{booktabs}

\usepackage{algorithmic}
\usepackage[ruled,vlined]{algorithm2e}

\usepackage{float}
\usepackage{url}
\usepackage{hyperref}
\usepackage[normalem]{ulem}

\def\BibTeX{{\rm B\kern-.05em{\sc i\kern-.025em b}\kern-.08em
    T\kern-.1667em\lower.7ex\hbox{E}\kern-.125emX}}
\begin{document}

\title{A Gradient-Regularized Deep Learning Method for Restoring Image in Myopic Vision\\
}

\author{
  \makebox[\textwidth][s]{
    \small
    \begin{tabular}{c}
      1\textsuperscript{st} Kebin Contreras \\
      \textit{Physics School} \\
      \textit{Univ. Industrial of Santander}\\
      Bucaramanga, Colombia
    \end{tabular}
    \hfill
    \begin{tabular}{c}
      2\textsuperscript{nd} Luis Toscano-Palomino \\
      \textit{Dept. of Computer Science} \\
      \textit{Univ. Industrial of Santander}\\
      Bucaramanga, Colombia
    \end{tabular}
    \hfill
    \begin{tabular}{c}
      3\textsuperscript{rd} Rafael Torres \\
      \textit{Physics School} \\
      \textit{Univ. Industrial of Santander}\\
      Bucaramanga, Colombia
    \end{tabular}
    \hfill
    \begin{tabular}{c}
      4\textsuperscript{th} Jorge Bacca \\
      \textit{Dept. of Computer Science} \\
      \textit{Univ. Industrial of Santander}\\
      Bucaramanga, Colombia
    \end{tabular}
  } \\[0.2cm]
}

\maketitle

\begin{abstract}
Myopic vision degrades the quality of human vision due to spherical eye defocus aberrations that impair the perception of fine spatial details. To address this limitation, this work proposes a deep learning-based approach to computationally design transformed images that allow individuals with myopia in fixed distance to perceive scenes with enhanced clarity. The method trains a neural network to counteract defocus aberrations modeled using the second-order Zernike polynomial \( Z_2^0 \). The approach employs a U-Net architecture trained with a loss function combining reconstruction error minimization with spatial difference structural preservation through $\ell_1$ regularization. Training is carried out exclusively on the KITTI dataset, while testing is performed on DIV2K and ImageNet datasets, demonstrating generalization across diverse image distributions. Under myopic conditions modeled by \( Z_2^0 \), baseline performance shows visual degradation with PSNR values ranging from 19.47 to 27.42 dB across datasets and severity levels. The proposed gradient-regularized method achieves consistent improvements, delivering PSNR gains ranging from 1.48 to 7.34 dB and significantly lower perceptual compared to classical baselines.
\end{abstract}


\section{Introduction}
\label{sec:intro}

\begin{figure*}[t]
    \centering
    \includegraphics[width=1.0\linewidth]{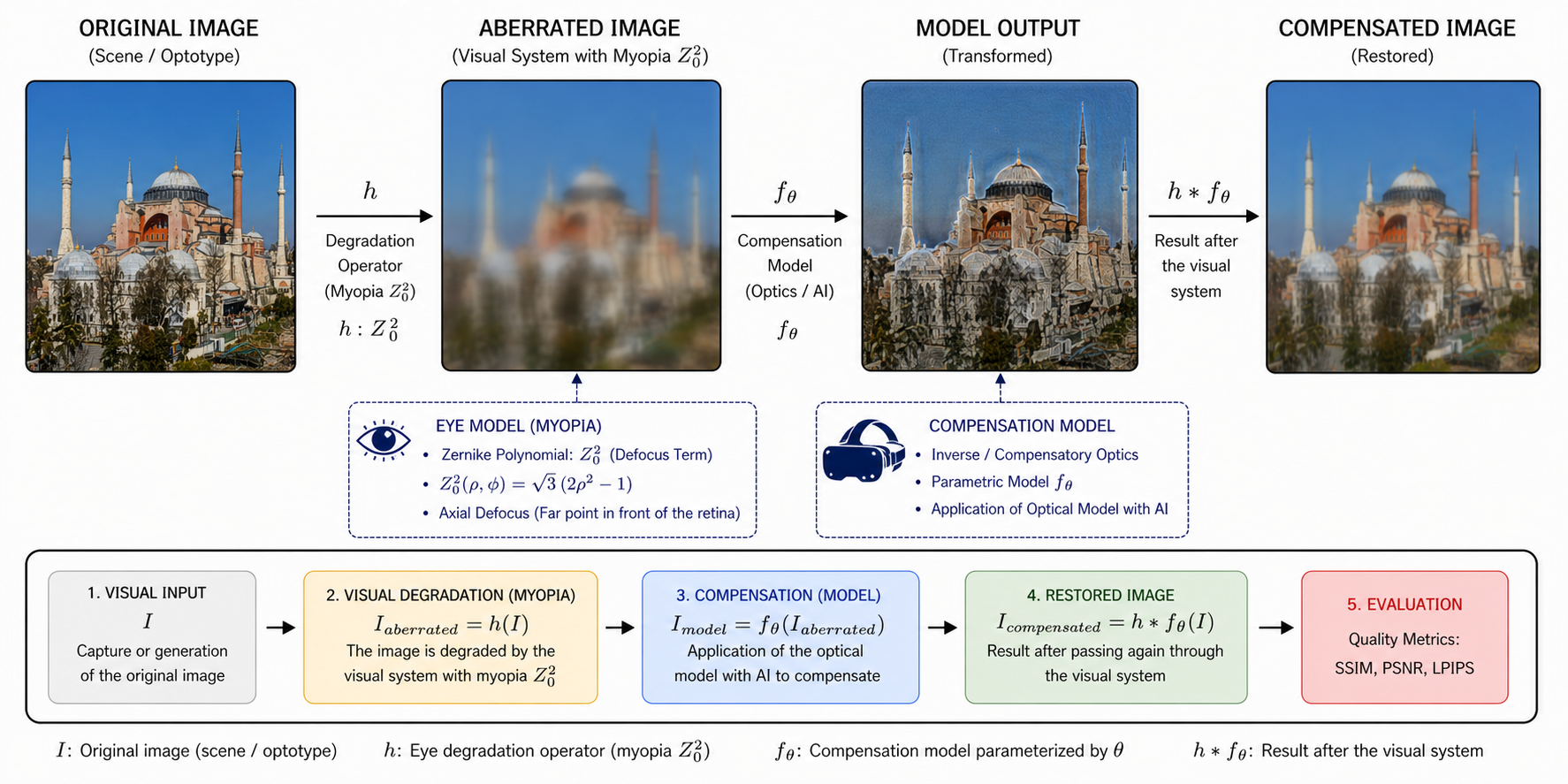} 
    \caption{Proposed computational pre-compensation pipeline. The process involves modeling the myopic visual system using Zernike polynomials ($Z_2^0$), applying a deep learning-based compensation model ($f_\theta$), and evaluating the restored result ($h * f_\theta$) through signal and perceptual metrics.}
    \label{fig:method_proposed}
\end{figure*}

Visual quality in human vision is essential for everyday tasks, as even minor eye degradations can affect perception and decision-making \cite{monroy2025autoregressive}. Among refractive errors, myopia (nearsightedness) represents one of the most prevalent visual impairments worldwide, affecting more than 2.6 billion people and projected to affect nearly half of the world population by 2050 \cite{holden2016global}. Myopic aberrations degrade the quality of visual input by causing blurring of distant objects, distorting spatial information, and reducing the ability to resolve fine details \cite{demir2015}. These imperfections impact daily activities that require distance vision, such as driving, sports, and classroom learning \cite{demir2015, badescu2015, contreras2025high}.

To quantitatively model myopic aberrations in the human eye, Zernike polynomials serve as mathematical tools to represent the aberrated wavefront emerging at the pupil plane after light passes through the eye’s optics \cite{noll1976}. These polynomials decompose the wavefront distortions into a series of orthogonal basis functions on a unit disk, providing a compact description of ocular imperfections \cite{badescu2015}. Specifically, the second-order Zernike polynomial \(Z_2^0\) represents defocus aberrations corresponding to spherical refractive errors that cause the uniform blurring observed in myopia \cite{mahajan1981}.

Historically, myopia correction has been based on physical optical elements such as concave lenses with specific diopter strengths tailored to individual prescriptions \cite{sundy2016}. However, traditional corrective approaches face several limitations including physical constraints, comfort issues, and challenges in dynamic correction for varying visual demands \cite{demir2015}. Moreover, the social and psychological implications of visual impairment can further limit the acceptance of conventional treatments \cite{demir2015}. Computational strategies have emerged as promising alternatives, proposing digital modification of visual content to compensate for optical aberrations.
Similarly, \cite{huang2014} proposed a light field-based eyeglasses-free display; however, their approach exhibits limited generalization since the hardware-dependent calibration fails under changing user head positions, off-axis viewing angles, or varying screen distances, and it does not rely on a formal optical aberration model grounded in wavefront theory.

This work addresses these limitations by proposing a gradient-regularized computational strategy specifically designed for myopia pre-compensation. As illustrated in Fig. \ref{fig:method_proposed}, our approach models the eye's degradation using a physics-based point spread function (PSF) derived from the second-order Zernike polynomial \(Z_2^0\) to accurately simulate spherical defocus \cite{watson2015}. Unlike conventional image restoration designed for digital sensors, we focus on scenarios with fixed viewing distances, such as digital displays, where a spatially invariant model provides an appropriate first-order approximation. Our method pre-transforms content ($f_\theta$) so that it appears clear to a myopic human observer after passing through their aberrated optics ($h$). By centering the study on this specific refractive error, we establish a robust baseline for computational correction before extending to more complex aberrations such as astigmatism. 

The proposed approach trains a U-Net architecture \cite{guan2019} on the KITTI dataset \cite{geiger2012} using a gradient-regularized loss that balances pixel fidelity with structural preservation through $\ell_1$-norm constraints on spatial gradients \cite{perona1990}. As shown in the final stages of Fig. \ref{fig:method_proposed}, the resulting model demonstrates strong generalization on the DIV2K \cite{agustsson2017} and ImageNet \cite{deng2009} datasets, outperforming classical Wiener filtering and non-regularized baselines in both signal fidelity (PSNR/SSIM) and perceptual metrics (LPIPS).

\section{Modeling Myopic Defocus Using Zernike Polynomials}

In practice, the severity of blur perceived by a myopic individual depends on both the optical aberration and the distance of objects in the scene. However, in the context of fixed-distance displays (e.g., monitors, AR/VR headsets, or handheld devices), all content is projected onto a plane at a constant distance from the eye. Under this condition, spherical defocus caused by myopia manifests as a spatially invariant blur across the image plane. For this reason, we adopt a PSF derived from the Zernike polynomial $Z_2^0$ as a tractable and widely used approximation of myopic vision~\cite{noll1976,mahajan1981}. For myopia modeling, the second-order Zernike polynomial \(Z_2^0\) represents pure aberrations of defocus. This mode corresponds to spherical wavefront errors and is mathematically expressed as:
\begin{equation}
Z_2^0(\rho, \theta) = \sqrt{3}(2\rho^2 - 1)
\end{equation}
where \((\rho, \theta)\)~\cite{noll1976} are the polar coordinates. The \(Z_2^0\) polynomial captures the characteristic of myopic vision: uniform spherical defocus that affects all spatial frequencies equally, resulting in rotationally symmetric blurring throughout the visual field. This distinguishes myopia from other aberrations, such as astigmatism, which introduce directional distortions~\cite{watson2015}.
For an eye affected by myopia, the incoming wavefront can be modeled as:
\begin{equation}
W(\rho, \theta) = \alpha Z_2^0(\rho, \theta)
\end{equation}
where the coefficient \( \alpha \) determines the severity of myopic defocus. Positive values typically correspond to myopic conditions, with larger magnitudes indicating more severe myopia requiring stronger corrective power~\cite{mahajan1981}.

The complex pupil function that describes light propagation through the myopic eye is given by:
\begin{equation}
P_{\lambda}(\rho, \theta) = A(\rho, \theta) \exp\left( i \frac{2\pi}{\lambda} W(\rho, \theta) \right)
\end{equation}
where \( A(\rho, \theta) \) represents the amplitude transmission function of the pupil and \( \lambda \) is the wavelength of light. For myopic modeling, it is typically assumed that the amplitude function is uniform across the pupil aperture~\cite{watson2015}.

The PSF characterizing the myopic eye's imaging properties is obtained through Fourier optics principles. The PSF represents the intensity distribution in the image plane and is computed as follows:
\begin{equation}
h_{\lambda}(x, y) = \left| \mathcal{F}_{\rho, \theta \rightarrow x, y} \left\{ P_{\lambda}(\rho, \theta) \right\} \right|^2,
\end{equation}
where \( \mathcal{F} \) denotes the two-dimensional Fourier transform, and the transformation from polar to Cartesian coordinates \((x = \rho \cos(\theta), y = \rho \sin(\theta))\) is applied for spatial domain representation. The PSF maintains rotational symmetry because the Zernike polynomial \(Z_2^0\) has angular order \(m=0\), indicating no angular dependence on \(\theta\). This results in a circularly symmetric point spread function where the blur kernel exhibits uniform characteristics in all radial directions, which accurately represents the isotropic nature of spherical defocus aberrations in myopic vision. All spatial frequencies experience similar degradation, which is characteristic of spherical aberrations and distinguishes defocus from other optical imperfections that may affect different frequency bands unequally.

The degraded image observed by a myopic individual is modeled as the convolution of the original scene with the myopic PSF:
\begin{equation}
I_{\text{myopic}}^\lambda(x, y) = h(x, y)_\lambda * I_\lambda
\end{equation}
where \( * \) denotes the convolution operation. This mathematical framework provides the foundation for accurately simulating myopic vision and developing computational correction strategies based on optical physics~\cite{silva2020a,silva2020b}.

\begin{table*}[h]
\centering
\caption{\small Comparative quantitative evaluation of image restoration across datasets under varying myopic defocus severities ($S$). Values represent Mean $\pm$ Standard Deviation. The best results are highlighted in bold, while the second-best are \underline{underlined}, arrows ($\uparrow$ and $\downarrow$).} 
\label{tab:cross_domain_results}
\vspace{-0.5em}
\resizebox{\textwidth}{!}{ 
\setlength{\tabcolsep}{4pt} 
\renewcommand{\arraystretch}{1.1}
\begin{tabular}{l ccc ccc ccc}
\toprule
\multirow{2}{*}{\textbf{Method / Severity (S)}} & \multicolumn{3}{c}{\textbf{KITTI \cite{geiger2012}}} & \multicolumn{3}{c}{\textbf{DIV2K \cite{agustsson2017}}} & \multicolumn{3}{c}{\textbf{ImageNet \cite{deng2009}}} \\
\cmidrule(lr){2-4} \cmidrule(lr){5-7} \cmidrule(lr){8-10}
& SSIM$\uparrow$ & PSNR$\uparrow$ & LPIPS$\downarrow$ & SSIM$\uparrow$ & PSNR$\uparrow$ & LPIPS$\downarrow$ & SSIM$\uparrow$ & PSNR$\uparrow$ & LPIPS$\downarrow$ \\
\midrule

\multicolumn{10}{l}{\textbf{S = 0.50 (Mild Aberration)}} \\
\midrule
Baseline (Uncorrected) & 0.848 $\pm$ 0.047 & 26.10 $\pm$ 1.43 & 0.237 $\pm$ 0.027 & 0.674 $\pm$ 0.115 & 23.31 $\pm$ 2.64 & 0.386 $\pm$ 0.084 & 0.840 $\pm$ 0.076 & 27.42 $\pm$ 3.29 & 0.237 $\pm$ 0.069 \\ 
Wiener Deconvolution   & 0.895 $\pm$ 0.032 & 26.96 $\pm$ 1.47 & 0.080 $\pm$ 0.018 & 0.746 $\pm$ 0.092 & 23.77 $\pm$ 2.80 & 0.304 $\pm$ 0.093 & 0.882 $\pm$ 0.057 & 28.26 $\pm$ 3.42 & 0.105 $\pm$ 0.047 \\
UNet (MSE Loss)        & \underline{0.989 $\pm$ 0.003} & \textbf{34.73 $\pm$ 0.78} & \underline{0.034 $\pm$ 0.009} & \underline{0.897 $\pm$ 0.056} & \textbf{28.29 $\pm$ 3.21} & \underline{0.220 $\pm$ 0.083} & \textbf{0.973 $\pm$ 0.024} & \textbf{34.61 $\pm$ 3.12} & \underline{0.056 $\pm$ 0.035} \\
\textbf{Ours (Unet + Reg)} & \textbf{0.993 $\pm$ 0.003} & \underline{33.44 $\pm$ 1.00} & \textbf{0.023 $\pm$ 0.009} & \textbf{0.900 $\pm$ 0.065} & \underline{27.79 $\pm$ 3.38} & \textbf{0.183 $\pm$ 0.078} & \underline{0.968 $\pm$ 0.037} & \underline{34.60 $\pm$ 4.36} & \textbf{0.041 $\pm$ 0.031} \\
\midrule

\multicolumn{10}{l}{\textbf{S = 1.00 (Moderate Aberration)}} \\
\midrule
Baseline (Uncorrected) & 0.750 $\pm$ 0.072 & 24.05 $\pm$ 1.35 & 0.440 $\pm$ 0.050 & 0.572 $\pm$ 0.145 & 22.02 $\pm$ 2.63 & 0.515 $\pm$ 0.099 & 0.764 $\pm$ 0.107 & 25.71 $\pm$ 3.30 & 0.395 $\pm$ 0.098 \\ 
Wiener Deconvolution   & 0.803 $\pm$ 0.055 & 22.05 $\pm$ 1.47 &0.218 $\pm$ 0.029 & 0.633 $\pm$ 0.133 & 20.78 $\pm$ 2.88 & 0.341 $\pm$ 0.087 & 0.802 $\pm$ 0.100 & 24.40 $\pm$ 3.77 & \underline{0.188 $\pm$ 0.064} \\
UNet (MSE Loss)  & \underline{0.894 $\pm$ 0.033} & \textbf{28.02 $\pm$ 1.35} & \textbf{0.155 $\pm$ 0.036} & \underline{0.781 $\pm$ 0.096} & \textbf{25.16 $\pm$ 3.24} & \underline{0.281 $\pm$ 0.093} & \underline{0.898 $\pm$ 0.058} & \textbf{30.24 $\pm$ 4.28} & \textbf{0.127 $\pm$ 0.062} \\
\textbf{Ours (Unet + Reg)} & \textbf{0.911 $\pm$ 0.032} & \underline{27.28 $\pm$ 1.27} &  \underline{0.158 $\pm$ 0.035} & \textbf{0.798 $\pm$ 0.098} & \underline{24.64 $\pm$ 3.16} & \textbf{0.272 $\pm$ 0.098} & \textbf{0.906 $\pm$ 0.062} & \underline{29.48 $\pm$ 4.20} & \textbf{0.127 $\pm$ 0.064} \\
\midrule

\multicolumn{10}{l}{\textbf{S = 2.00 (Severe Aberration)}} \\
\midrule
Baseline (Uncorrected) & 0.665 $\pm$ 0.090 & 20.92 $\pm$ 1.36 & 0.489 $\pm$ 0.044 & 0.511 $\pm$ 0.148 & 20.32 $\pm$ 2.43 & 0.502 $\pm$ 0.089 & 0.698 $\pm$ 0.123 & 23.08 $\pm$ 3.08 & 0.423 $\pm$ 0.082 \\ 
Wiener Deconvolution   & 0.714 $\pm$ 0.066 & 18.25 $\pm$ 1.27 &0.283 $\pm$ 0.028 & 0.590 $\pm$ 0.131 & 19.00 $\pm$ 2.55 & 0.369 $\pm$ 0.079 & 0.740 $\pm$ 0.119 & 21.18 $\pm$ 3.37 & 0.229 $\pm$ 0.069\\

UNet (MSE Loss)        & \underline{0.816 $\pm$ 0.052} & \textbf{22.96 $\pm$ 1.41} &  \underline{0.221 $\pm$ 0.023} & \underline{0.724 $\pm$ 0.103} & \textbf{22.78 $\pm$ 2.85} & \underline{0.289 $\pm$ 0.071} & \underline{0.833 $\pm$ 0.084} & \textbf{26.00 $\pm$ 3.69} & \underline{0.176 $\pm$ 0.056} \\

\textbf{Ours (Unet + Reg)} & \textbf{0.848 $\pm$ 0.048} & \underline{22.68 $\pm$ 1.41} & \textbf{0.202 $\pm$ 0.033} & \textbf{0.764 $\pm$ 0.100} & \underline{22.47 $\pm$ 2.85} & \textbf{0.253 $\pm$ 0.077} & \textbf{0.859 $\pm$ 0.083} & \underline{25.79 $\pm$ 3.79} & \textbf{0.151 $\pm$ 0.060} \\
\midrule

\multicolumn{10}{l}{\textbf{S = 3.00 (Extreme Aberration)}} \\
\midrule

Baseline (Uncorrected) & 0.646 $\pm$ 0.092 & 19.55 $\pm$ 1.34 & 0.482 $\pm$ 0.037 & 0.496 $\pm$ 0.149 & 19.47 $\pm$ 2.34 & 0.477 $\pm$ 0.079 & 0.680 $\pm$ 0.127 & 21.77 $\pm$ 2.99 & 0.417 $\pm$ 0.071 \\

Wiener Deconvolution & 0.695 $\pm$ 0.069 & 17.32 $\pm$ 1.19 & 0.302 $\pm$ 0.029 & 0.581 $\pm$ 0.132 & 18.65 $\pm$ 2.45 & 0.371 $\pm$ 0.076 & 0.725 $\pm$ 0.123 & 20.40 $\pm$ 3.24 & 0.238 $\pm$ 0.067 \\

UNet (MSE Loss) &
\underline{0.797 $\pm$ 0.060} &
\textbf{21.21 $\pm$ 1.42} &
\underline{0.237 $\pm$ 0.025} &
\underline{0.702 $\pm$ 0.108} &
\textbf{21.78 $\pm$ 2.77} &
\underline{0.302 $\pm$ 0.070} &
\underline{0.820 $\pm$ 0.089} &
\textbf{24.42 $\pm$ 3.58} &
\underline{0.188 $\pm$ 0.058} \\

\textbf{Ours (Unet + Reg)} &
\textbf{0.827 $\pm$ 0.051} &
\underline{21.03 $\pm$ 1.39} &
\textbf{0.221 $\pm$ 0.031} &
\textbf{0.753 $\pm$ 0.098} &
\underline{21.35 $\pm$ 2.64} &
\textbf{0.249 $\pm$ 0.067} &
\textbf{0.846 $\pm$ 0.080} &
\underline{23.80 $\pm$ 3.21} &
\textbf{0.168 $\pm$ 0.057} \\

\bottomrule
\end{tabular}
} 
\vspace{-0.5em}
\end{table*}
\section{Gradient-Regularized Image Correction}

The proposed computational correction strategy aims to generate transformed images that, when viewed by people with myopia, appear sharp and clear. Although the optical degradation is wavelength-dependent as shown in Eq. (5), digital images in this framework are processed using three independent color bands (RGB). For notational simplicity in the subsequent formulation, the wavelength subscript \( \lambda \) is omitted hereafter, with the understanding that operations are applied independently across all color channels. Let \( I \) represent the original sharp image. The objective is to find a transformed image \( I_{\text{transformed}} \) such that:
\begin{equation}
I \approx I_{\text{view}} = h * I_{\text{transformed}}.
\end{equation}

Therefore, this method seeks a transformation that compensates for the known degradation characteristics of myopic vision. Rather than restoring the scene at the optical level, the goal is to pre-process the image such that, when it is viewed through a myopic visual system, it appears perceptually sharp. This formulation allows the correction to be performed entirely in the digital domain, removing the dependence on physical lenses and enabling adaptable, content-aware correction. However, due to the ill-posed nature of deconvolution and the need to preserve fine structural details, a data-driven learning framework is employed. Specifically, we adopt a deep convolutional architecture to learn a transformation that counteracts the effects of the myopic PSF, while regularizing for perceptual consistency through spatial

A U-Net architecture~\cite{guan2019}, denoted as \( f_\theta \), is used to learn the transformation mapping \( I_{\text{transformed}} = f_\theta(I) \). The network is trained to generate corrected images that, when combined with the myopic point spread function \( h \), closely approximate the original sharp images.

A key component of our method is the preservation of high-frequency structural information during image reconstruction~\cite{bacca2025}. To achieve this, we introduce a gradient regularization term that penalizes inconsistencies in edges and textures, guiding the network to maintain sharpness and avoid overly smoothed results. The optimization problem is formulated as \vspace{-0.5em}
\begin{align}
& \theta^* = \arg\min_{\theta} \bigg[
\mathbb{E}_I \left\| h * f_\theta(I) - I \right\|_2^2 
+ \\  &  \alpha \big( \left\| \nabla_x (h * f_\theta(I)) - \nabla_x I \right\|_1 
+ \left\| \nabla_y (h * f_\theta(I))  - \nabla_y I \right\|_1 \big) \bigg], \nonumber
\label{eq:cross_domain_optimization}
\end{align}
where the first term enforces perceptual similarity of the corrected image $I_{view}=h * f_\theta(I)$, and the second term enforces consistency of horizontal and vertical gradients~\cite{perona1990} controlled with the hyperparameter $\alpha$.

This gradient constraint plays a central role in ensuring cross-domain generalization. By minimizing $\ell_1$-nom 
 differences between gradients, the model preserves sharp transitions and structural coherence across diverse visual content, making the correction transferable to image types not seen during training.  The use of the $\ell_1$-norm promotes sparsity in gradient differences, effectively maintaining edges while allowing smooth regions, thus compensating for optical degradation without sacrificing perceptual quality.

The training procedure, described in Algorithm~\ref{alg:cross_domain_algorithm}, learns a transformation that compensates for myopic degradation by leveraging a physics-informed framework enhanced with gradient-based regularization. The model is designed to generalize across diverse visual domains, ensuring applicability to a wide range of image types regardless of scene characteristics or dataset origin.

\begin{algorithm}[ht]
\caption{\small  Gradient-Regularized Training for Myopia Correction}
\label{alg:cross_domain_algorithm}
\SetKwInOut{Input}{Input}
\SetKwInOut{Output}{Output}
\Input{$\mathcal{D}_{\text{KITTI}} = \{I_i\}_{i=1}^M$, $c_{2,0}$, $\lambda$, $\theta$, $\eta$}
\ForEach{batch of images $I_i \in \mathcal{D}_{\text{KITTI}}$}{
    $W \leftarrow \alpha Z_2^0$ \tcp*{Myopic wavefront}
    $P \leftarrow A \exp\left(i \frac{2\pi}{\lambda} W\right)$ \tcp*{Pupil function}
    $h \leftarrow |\mathcal{F}\{P\}|^2$ \tcp*{Myopic PSF}
    $I_{\text{transformed}} \leftarrow f_\theta(I_i)$ \tcp*{Network prediction}
    $\mathcal{L}_{\text{recon}} \leftarrow \|h * I_{\text{transformed}} - I_i\|_2^2$ \tcp*{Recon. loss}
    $\mathcal{L}_{\text{grad}} \leftarrow \|\nabla_x(h * I_{\text{transformed}}) - \nabla_x I_i\|_1 + \|\nabla_y(h * I_{\text{transformed}}) - \nabla_y I_i\|_1$ \tcp*{Grad. loss}
    $\mathcal{L}_{\text{total}} \leftarrow \mathcal{L}_{\text{recon}} + \alpha \mathcal{L}_{\text{grad}}$ \tcp*{Total loss}
    $\theta \leftarrow \theta - \eta \nabla_\theta \mathcal{L}_{\text{total}}$ \tcp*{Update params}
}
\Output{Optimized parameters $\theta^*$}
\end{algorithm}

\begin{figure}[h]
    \centering \vspace{-1em}
    \includegraphics[width=1\linewidth]{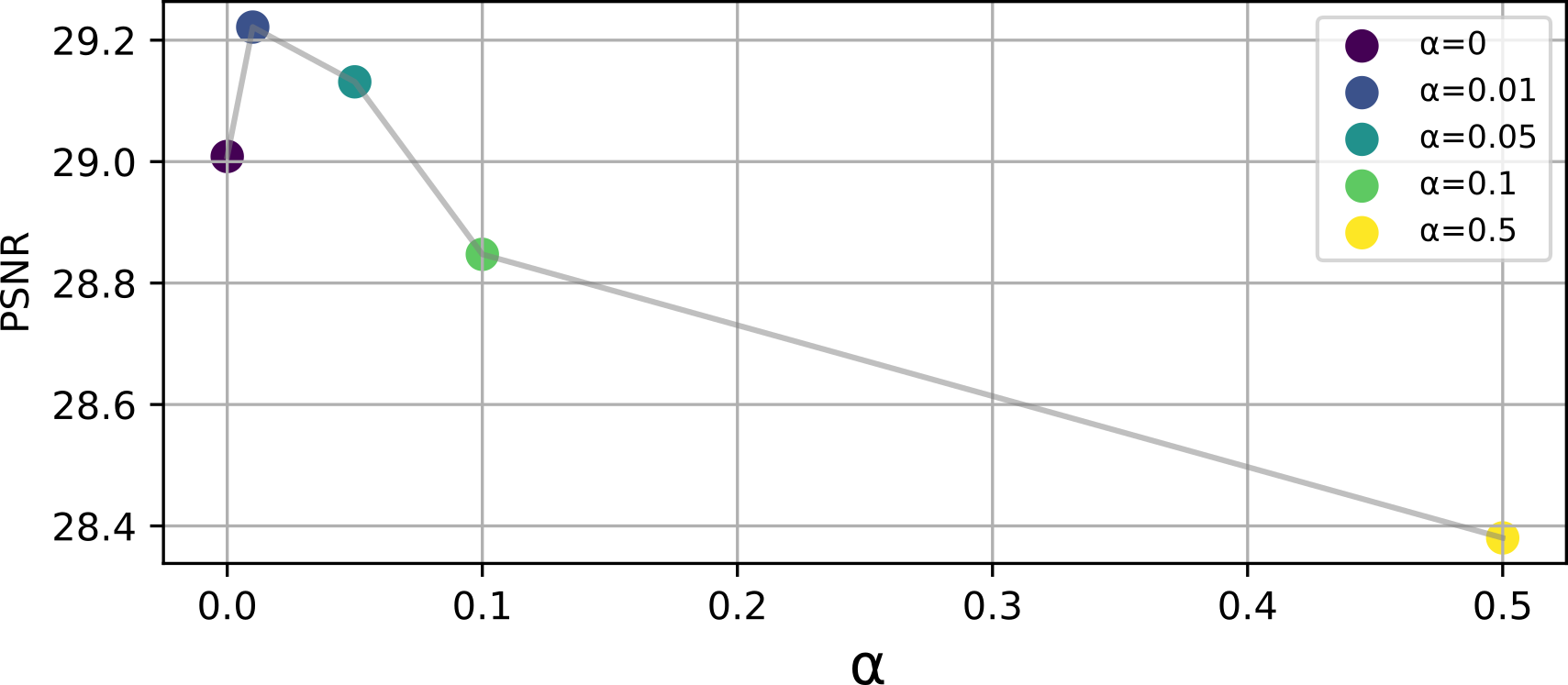}
    \vspace{-2em}
    \caption{\small Hyperparameter tuning for the gradient regularization weight \(\alpha\). The PSNR is plotted for different values of \(\alpha\), showing that \(\alpha = 0.01\) achieves the best performance.}
    \label{fig:lambda-ablation}
\end{figure}

\begin{figure*}[h]
    \centering
    \includegraphics[width=1\linewidth]{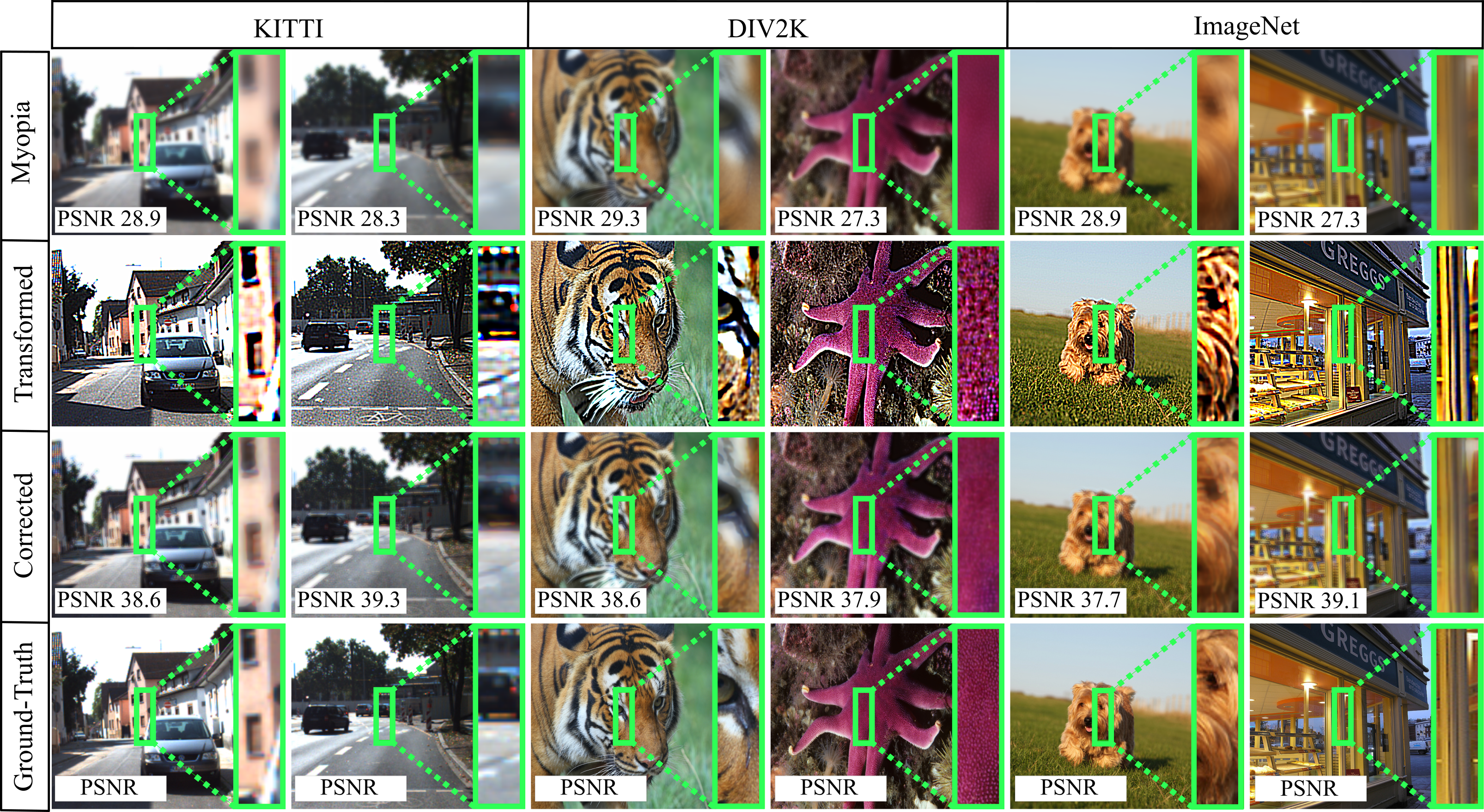} 
    \caption{\small Cross-domain visual results comparing myopic with severity level 2.0 degradation and gradient-regularized correction. PSNR values are shown for KITTI (training), DIV2K (natural), and ImageNet (object-centric) datasets.} \vspace{-1em}

    \label{fig:cross_domain_visual}
\end{figure*}

\vspace{-0.1cm}

\section{SIMULATIONS AND RESULTS}

The proposed method is evaluated through a cross-domain strategy to assess reconstruction accuracy and generalization. A U-Net architecture is trained exclusively on KITTI~\cite{geiger2012}. Hyperparameter tuning was performed via grid search for \( \alpha \in \{0, 0.01, 0.05, 0.1, 0.5\} \), showing that \( \alpha = 0.01 \) for the Sobel gradient regularization provides the optimal balance between pixel fidelity and structural sharpness (Fig.~\ref{fig:lambda-ablation}). Myopic defocus is modeled via the \(Z_2^0\) Zernike polynomial across four severity levels. Furthermore, the mean and standard deviation reported in Table~\ref{tab:cross_domain_results} were computed across 100 test images for each dataset, capturing the variance of the model's performance across diverse visual content.

Performance is evaluated using Peak Signal-to-Noise Ratio (PSNR), Structural Similarity Index (SSIM), and Learned Perceptual Image Patch Similarity (LPIPS). While PSNR and SSIM measure fidelity and structural consistency, LPIPS evaluates perceptual similarity in deep feature space.

Table~\ref{tab:cross_domain_results} compares the proposed method against classical Wiener deconvolution~\cite{wiener1949} and a UNet trained with MSE loss across the multiple myopic defocus severities. The proposed gradient-regularized model consistently achieves the best or second-best LPIPS values across all datasets, indicating superior perceptual quality.

At mild degradation ($S=0.5$), the model obtains the best LPIPS in all domains while maintaining PSNR values close to the UNet-MSE baseline. As degradation severity increases, the advantage of gradient regularization becomes more evident, preserving sharper contours and structural details compared to Wiener deconvolution, which introduces ringing artifacts, and MSE-based UNet, which tends to over-smooth textures.


Under severe degradations ($S=2.0$ and $S=3.0$), the proposed method consistently surpasses Wiener filtering and remains competitive with UNet-MSE, demonstrating stronger perceptual robustness and better cross-domain generalization. Qualitative results in Fig.~\ref{fig:cross_domain_visual} further validate the recovery of fine textures and edge consistency. In particular, the \emph{Myopia} images correspond to the degraded observation $h \ast I$, the \emph{Transformed} images denote the network output $f_\theta(I)$, and the \emph{Corrected} images correspond to the perceived result after passing through the myopic visual system, $h \ast f_\theta(I)$, which is compared with the \emph{Ground Truth} image $I$.

\section{CONCLUSIONS} 

This work presents a novel gradient-regularized computational framework for myopic vision pre-compensation that effectively bridges physics-based optical modeling with deep learning. The proposed method demonstrates that integrating the second-order Zernike polynomial $Z_2^0$ into a U-Net training pipeline allows for the precise inversion of spherical defocus aberrations. Quantitatively, the approach achieves consistent PSNR gains across diverse distributions, ranging from 1.48 dB to 7.34 dB. 

The primary technical contribution lies in the adoption of a gradient-regularized loss function, which prioritizes structural coherence over simple pixel-wise error minimization. This strategy successfully mitigates the high-frequency artifacts and ringing effects typically associated with classical Wiener deconvolution, as evidenced by the significantly lower LPIPS scores achieved across all datasets. Furthermore, the model's robust performance on DIV2K and ImageNet, despite being trained exclusively on KITTI, provides empirical evidence of its cross-domain generalization capabilities. These findings validate the use of perceptual metrics as a reliable proxy for human visual judgment in the absence of clinical trials

\bibliographystyle{IEEEtran}
\bibliography{referencias}

@article{holden2016global,
  author = {B. A. Holden and T. R. Fricke and D. A. Wilson and M. Jong and K. S. Naidoo and P. Sankaridurg and T. Y. Wong and T. J. Naduvilath and S. Resnikoff},
  title = {Global prevalence of myopia and high myopia and temporal trends from 2000 through 2050},
  journal = {Ophthalmology},
  volume = {123},
  number = {5},
  pages = {1036--1042},
  year = {2016}
}

@article{badescu2015,
  author = {S. V. B\u{a}descu and R. Barac and S. Schmitzer and C. T\u{a}taru},
  title = {The influence of optical aberrations in refractive surgery},
  journal = {Romanian Journal of Ophthalmology},
  volume = {59},
  number = {4},
  pages = {217--222},
  year = {2015}
}

@article{demir2015,
  author = {M. Demir and S. A. Kurna and T. Sengor and T. G. Atakan and T. Sahin},
  title = {Assessment of aberrations and visual quality differences between myopic and astigmatic eyes before and after contact lens application},
  journal = {Northern Clinics of Istanbul},
  volume = {2},
  number = {1},
  pages = {1--6},
  year = {2015},
  month = {April},
  doi = {10.14744/nci.2015.87487}
}

@misc{monroy2025autoregressive,
  author = {B. Monroy and K. Contreras and J. Bacca},
  title = {Autoregressive high-order finite difference modulo imaging: high-dynamic range for computer vision applications},
  year = {2025},
  note = {[Online]. Available: https://arxiv.org/abs/2504.04228}
}

@misc{contreras2025high,
  author = {K. Contreras and B. Monroy and J. Bacca},
  title = {High dynamic range modulo imaging for robust object detection in autonomous driving},
  year = {2025},
  note = {[Online]. Available: https://arxiv.org/abs/2504.11472}
}

@article{sundy2016,
  author = {M. Sundy and D. McKnight and C. Eck and F. Rieger},
  title = {Visual acuity outcomes of toric lens implantation in patients undergoing cataract surgery at a residency training program},
  journal = {Missouri Medicine},
  volume = {113},
  number = {1},
  pages = {40--43},
  year = {2016}
}

@misc{bacca2025,
  author = {J. Bacca},
  title = {Projection-based correction for enhancing deep inverse networks},
  year = {2025},
  note = {[Online]. Available: https://arxiv.org/abs/2505.15777}
}

@article{huang2014,
  author = {F.-C. Huang and R. Raskar},
  title = {Eyeglasses-free display: towards correcting visual aberrations with computational light field displays},
  journal = {ACM Transactions on Graphics},
  volume = {33},
  number = {4},
  pages = {59:1--59:12},
  year = {2014},
  doi = {10.1145/2601097.2601130}
}

@article{noll1976,
  author = {R. J. Noll},
  title = {{Zernike} polynomials and atmospheric turbulence},
  journal = {Journal of the Optical Society of America},
  volume = {66},
  number = {3},
  pages = {207--211},
  year = {1976},
  doi = {10.1364/JOSA.66.000207}
}

@article{mahajan1981,
  author = {V. N. Mahajan},
  title = {{Zernike} circle polynomials and optical aberrations of systems with circular pupils},
  journal = {Journal of the Optical Society of America},
  volume = {71},
  number = {1},
  pages = {75--85},
  year = {1981},
  doi = {10.1364/JOSA.71.000075}
}

@article{watson2015,
  author = {A. B. Watson and J. D. Yellott and D. M. E. Cooper},
  title = {Computing human optical point spread functions},
  journal = {Journal of Vision},
  volume = {15},
  number = {2},
  pages = {26},
  year = {2015},
  doi = {10.1167/15.2.26}
}

@inproceedings{geiger2012,
  author = {A. Geiger and P. Lenz and R. Urtasun},
  title = {Are we ready for autonomous driving? the {KITTI} vision benchmark suite},
  booktitle = {Proceedings of the IEEE Conference on Computer Vision and Pattern Recognition},
  pages = {3354--3361},
  year = {2012}
}

@inproceedings{agustsson2017,
  author = {E. Agustsson and R. Timofte},
  title = {{NTIRE} 2017 challenge on single image super-resolution: dataset and study},
  booktitle = {Proceedings of the IEEE Conference on Computer Vision and Pattern Recognition Workshops},
  pages = {126--135},
  year = {2017}
}

@inproceedings{deng2009,
  author = {J. Deng and W. Dong and R. Socher and L.-J. Li and K. Li and L. Fei-Fei},
  title = {{ImageNet}: a large-scale hierarchical image database},
  booktitle = {Proceedings of the IEEE Conference on Computer Vision and Pattern Recognition},
  pages = {248--255},
  year = {2009}
}

@article{guan2019,
  author = {S. Guan and A. A. Khan and S. Sikdar and P. V. Chitnis},
  title = {Fully dense {UNet} for 2-D sparse photoacoustic tomography artifact removal},
  journal = {IEEE Journal of Biomedical and Health Informatics},
  volume = {24},
  number = {2},
  pages = {568--576},
  year = {2019}
}

@article{silva2020a,
  author = {A. Silva-Lora and R. Torres},
  title = {Superconical aplanatic ovoid singlet lenses},
  journal = {Journal of the Optical Society of America A},
  volume = {37},
  number = {7},
  pages = {1155--1165},
  year = {2020}
}

@article{silva2020b,
  author = {A. Silva-Lora and R. A. Torres},
  title = {Explicit Cartesian oval as a superconic surface for stigmatic imaging optical systems with real or virtual source or image},
  journal = {Proceedings of the Royal Society A},
  volume = {476},
  number = {2235},
  pages = {20190894},
  year = {2020}
}

@article{perona1990,
  author = {P. Perona and J. Malik},
  title = {Scale-space and edge detection using anisotropic diffusion},
  journal = {IEEE Transactions on Pattern Analysis and Machine Intelligence},
  volume = {12},
  number = {7},
  pages = {629--639},
  year = {1990}
}

@book{wiener1949,
  author    = {Norbert Wiener},
  title     = {Extrapolation, Interpolation, and Smoothing of Stationary Time Series},
  publisher = {MIT Press},
  address   = {Cambridge, Massachusetts},
  year      = {1949}
}

\end{document}